# A Brief Review of Explainable Artificial Intelligence in Healthcare

Zahra Sadeghi[1], Roohallah Alizadehsani[2,*], Mehmet Akif CIFCI[3], Samina Kausar[4], Rizwan Rehman[5], Priyakshi Mahanta[6], Pranjal Kumar Bora[7], Ammar Almasri[8], Rami S. Alkhawaldeh[9], Sadiq Hussain[10], Bilal Alatas[11], Afshin Shoeibi[12], Hossein Moosaei[13], Milan Hladík[14], Saeid Nahavandi[2,15], Panos M. Pardalos[16]

[1]Institute for Big Data Analytics, Faculty of Computer Science, Dalhousie University, Canada, zahra.sadeghi@dal.ca

[2]Institute for Intelligent Systems Research and Innovation (IISRI), Deakin University, Geelong, Australia.

[3]The Institute of Computer Technology, Tu Wien University, 1040 Vienna, Austria; mcifci@tuwien.ac.at

[4]University of Kotli Azad Jammu and Kashmir, Kotli Azad Kashmir, Pakistan,saminamalik7@yahoo.com

[5]Centre for Computer Science and Applications, Dibrugarh University, Assam, India,rizwan@dibru.ac.in

[6]Centre for Computer Science and Applications, Dibrugarh University, Assam, India,priyakshimahanta@dibru.ac.in

[7]Centre for Computer Science and Applications, Dibrugarh University, Assam, India, pranjalbora@dibru.ac.in

[8]Department of Management Information Sys, Al-Balqa Applied University, Salt 19117, Jordan, ammar.almasri@bau.edu.jo

[9]Department of Computer Information Systems, The University of Jordan, Aqaba 77110, Jordan, r.alkhawaldeh@ju.edu.jo

[10]Examination Branch, Dibrugarh University, Dibrugarh, Assam, India, sadiq@dibru.ac.in

[11]Department of Software Eng., Firat University, 23100 Elazig, Turkey, balatas@firat.edu.tr

[12]Data Science and Computational Intelligence Institute, University of Granada, Spain

[13]Department of Informatics, Faculty of Science, Jan Evangelista Purkyně University in Ústí nad Labem, Czech Republic

[14]Department of Applied Mathematics, School of Computer Science, Faculty of Mathematics and Physics, Charles University, Prague, Czech Republic

[15]Harvard Paulson School of Engineering and Applied Sciences, Harvard University, Allston, MA 02134, USA

[16]Center for Applied Optimization, Department of Industrial and Systems Engineering, University of Florida, Gainesville, 32611, USA

* Corresponding author email: r.alizadehsani@deakin.edu.au

**Abstract:**
XAI refers to the techniques and methods for building AI applications which assist end users to interpret output and predictions of AI models. Black box AI applications in high-stakes decision-making situations, such as medical domain have increased the demand for transparency and explainability since wrong predictions may have severe consequences. Model explainability and interpretability are vital successful deployment of AI models in healthcare practices. AI applications' underlying reasoning needs to be transparent to clinicians in order to gain their trust. This paper presents a systematic review of XAI aspects and challenges in the healthcare domain. The primary goals of this study are to review various XAI methods, their challenges, and related machine learning models in healthcare. The methods are discussed under six categories: Features-oriented methods, global methods, concept models, surrogate models, local pixel-based methods, and human-centric methods. Most importantly, the paper explores XAI role in healthcare problems to clarify its necessity in safety-critical applications. The paper intends to establish a comprehensive understanding of XAI-related applications in the healthcare field by reviewing the related experimental results. To facilitate future research for filling research gaps, the importance of XAI models from different viewpoints and their limitations are investigated.

I. **Introduction**

Explainable Artificial Intelligence (XAI) has become a significant topic in recent years because these models can make AI systems trustworthy, compliant, effective, and robust. XAI refers to the techniques and methods to build AI applications that end users can understand and interpret. The end users can be domain experts, data scientists, or even individuals without academic knowledge about AI. Great success of deep learning (DL) alongside its widespread deployment in real-world applications has ignited the desire for interpreting the rationale behind its decisions. Generally, users favour transparent AI models which can be interpreted or explained clearly. Before moving on, two conceptually different terms need to be clarified. Interpretability refers to providing human-understandable rules that define a system's decision-making mechanism. In comparison, explainability concerns creating a human-comprehensible interface for disentangling the internal AI decision-making function (1). The importance of AI explainability can be discussed from different viewpoints (1-4). First, explaining machine learning (ML) models is vital for verifying sensitive models such as those related to the human healthcare system. Medical experts need to ensure the models are trained correctly and the parameters on which they are dependent are consistent with their knowledge. For instance, if the post-hoc analysis results of an ML model conclude that sneezing is a sign of cancer, the medical doctor can immediately imply that the ML model is not trustworthy. Secondly, complex ML models such as deep neural networks are usually trained on very high-dimensional data and encapsulate salient features. Explaining these trained models will provide insightful information for experts in various fields of study such as Physics, Mathematics, and Chemistry. Using this information, scientists are able to discover new natural rules, obtain better observation about fundamental questions, and facilitate the advancement of research in these fields. Third, our everyday life is becoming more and more dependent on AI models in various senses. For example, many kinds of paperwork processing are handled by AI solutions and rejected applicants need to know the rejection reasons. Fourth, Neuroscience can benefit tremendously from AI Explainability to test and explain different hypotheses about the interaction between neurons in the brain and answer questions about computational activities and the learning mechanism of the brain.

AI explainability solutions based on post-hoc modelling, and analysis for ML models deciphering can be divided into model agnostic and model-specific methods. Model-agnostic approaches are general purpose and can be applied to almost all ML models regardless of their structure and training mechanism. One of the robust agnostic approaches is sensitivity analysis (SA) which attempts to reveal the contribution and impact of input factors on output prediction by changing input values and observing the amount of variation caused in the output (5). These methods indicate the sensitivity level of the output on each of the input variables based on different statistical features such as variance (6), derivative (7), and density (8). Sensitivity analysis can be applied globally or locally (9). Local SA methods rely on local perturbation of input values and measure the sensitivity based on the amount of variation of the output values. In the global SA techniques, the total possible values of input parameters are subject to change (10).

In contrast to model-agnostic approaches, model-specific methods can only be utilized for specific ML models. For example, many explainability mechanisms are developed to analyze trained deep neural networks. These approaches are known as deep network understanding and visualization (11). Activation Maximization (12), DeConvNet (13), inversion (14), deepDream (15), feature visualization analysis (16), and DeepLift (17) are some of the popular methods from this category which attempt to find the contribution of neurons of convolutional neural networks on their final decision through optimization and backpropagation. Moreover, specific approaches have been proposed for explaining Graph Neural Networks (18) and Recurrent Neural Networks (19).

Explainability solutions leverage various criteria such as trustworthiness, transferability, causality, and interactivity. Trustworthiness refers to the idea of seeking a simple model which makes an equal decision upon meeting a specific condition. Transferability explains the generalization capacity of a complex model and the reusable scenarios. In causality approach, the emphasis is towards finding cause and effect relationships (i.e. correlation) between variables. Some authors believe that for accurate interpretation of a model, the reasons behind its decisions must be uncovered. In this regard, counterfactual is a promising strategy to find the features contributing to a specific outcome (20). Another factor used in some AI explainability techniques is the model's ability to engage with end-users. Recent studies have focused on developing hybrid approaches which result in transparent models with representation power of existing black box architectures such as DL. Contextual Explanation Network (CEN) (21), Self-Explaining Neural Network (SENN) (22), BagNet (23), and TabNet (24) are typical examples of transparent models. The end-users' interest in understanding the reasons behind the decisions made by ML models demands further research. Model transparency is especially important in safety-critical applications such as medical domain. Therefore, our focus is on XAI in the healthcare domain and its challenges. The primary goals of our study are listed below:

(i) XAI methods identification and categorization
(ii) XAI literature review with special focus on healthcare domain
(iii) Ascertainment of XAI challenges and problems in healthcare.

The papers reviewed in this survey are selected using the keywords "Explainable AI" and "Interpretable Machine Learning", with a focus on "healthcare." The queries returned multitude of journal and conference articles. For the purpose of this survey, only peer-reviewed articles are considered for a systematic review. The Prisma model in Figure 1 demonstrates the overall papers reviewed in the study. Table 1 describes the explored databases. The bibliographic section of the articles was also scrutinized. The process was iterated until no more articles were found.

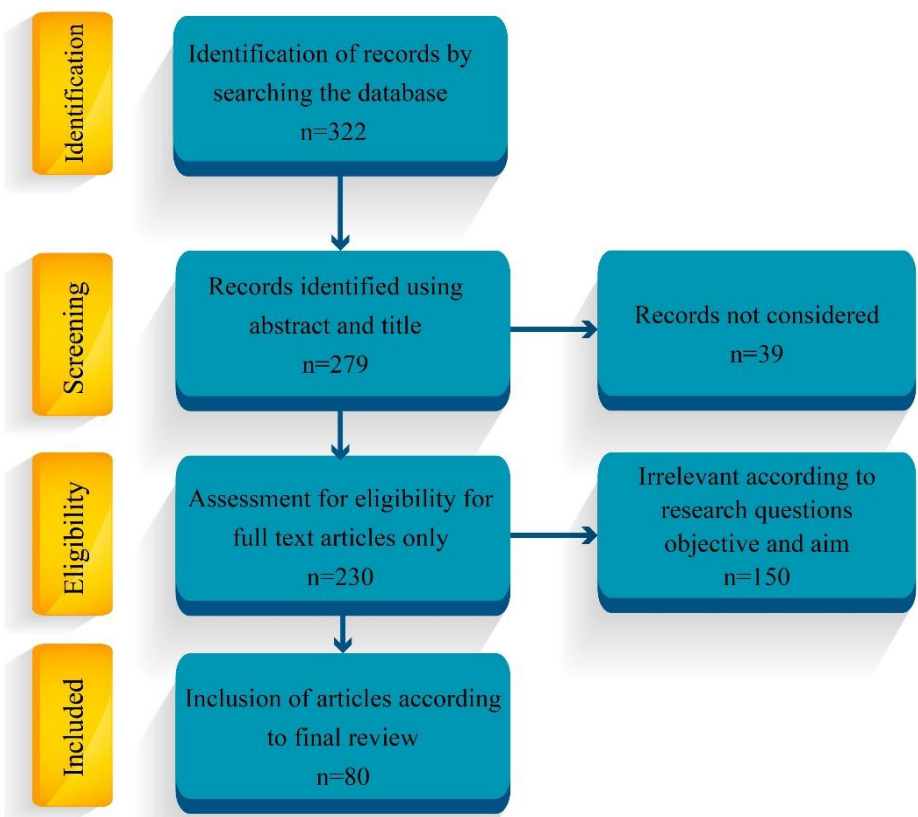

Figure 1. Prisma Model for the depiction of inclusion and exclusion of records

The procedure of detecting prospective research papers includes identification, screening, eligibility and inclusion of the chosen papers. A step-by-step flow chart of the detection procedure is illustrated in Figure 1. The utilized search engines were Google Scholar, Elsevier, Springer, etc. as depicted in

Table 1. The XAI papers irrelevant to health sector (39 articles) were not taken into account. A total of 279 articles were chosen for reviewing and after studying the titles and abstracts, 79 were excluded from the list. The preprint version and duplicate articles were also excluded. After evaluating the quality of published research works, 200 articles were selected 150 of which were excluded due to not being research articles or notions. Therefore, 50 articles were selected for a comprehensive review.

Table 1. Summary of search results and retrieved relevant articles

| Database Engines | Source Address | Number of search results | Number of relevant articles |
| --- | --- | --- | --- |
| Elsevier | https://www.elsevier.com | 1500 | 20 |
| Springer | https://www.springer.com | 1200 | 10 |
| Taylor & Francis | https://taylorandfrancis.com | 800 | 5 |
| Semantic Scholar | https://www.sematicscholar.org | 500 | 10 |
| ACM Digital Library | https://www.acm.org | 1000 | 10 |
| IEEE Xplore | https://ieeexplore.ieee.org | 2000 | 15 |

Table 1 shows the search results for relevant articles in various academic database search engines. The number of search results indicates how many articles were returned by the search engines for the given keywords. The number of relevant articles refers to the number of articles that passed the initial screening process and were deemed potentially relevant for a comprehensive review. IEEE Xplore database engine returned the highest number of search results (2000), while Taylor & Francis returned the fewest (800). However, the number of relevant articles is not necessarily proportional to the number of search results, as some articles may be excluded during the screening process. For example, Elsevier returned the highest number of relevant articles (20) despite not having the highest number of search results.

XAI sheds light on black box ML models to aid with understanding the logic behind their decision making. Black box models may not even be explainable by their designers (25). Explainability is an influential tool for justifying AI based decisions. It can assist to validate predictions, for enhancing models and for gaining new insights into the problem at hand that leads towards more trustworthy AI systems.

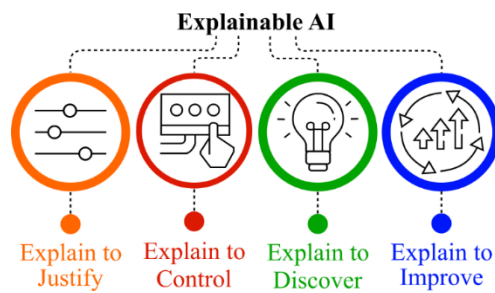

Figure 2. Motivations for XAI

Motivations for implementing XAI systems are graphically depicted in Figure 2 (25). As can be seen, there are several key drivers for XAI development such as increasing model transparency, improving accountability, and enhancing trust in AI systems. Other important drivers include the need for regulatory compliance, the desire for more effective decision-making, and the importance of ethical considerations in AI development (26).

The rest of the paper is organized as follows. Section 2 describes XAI methods. Section3 depicts XAI for decision makers. Applications of XAI in healthcare are showcased in Section 4. Section 5 describes the challenge of interpretability in healthcare while section 6 is the paper conclusion.

II. **Explainable Artificial Intelligence Methods**

Research in XAI can be categorized into six main groups: Feature-oriented methods, global methods, concept models, surrogate models, local pixel-based methods, and human-centric methods.

**2.1 Feature-oriented methods**

Shapley Additive exPlanation (SHAP) employs game theory to explain the outcomes of ML techniques. For each sample $x = [x_1, ..., x_n]$, contribution of each feature $x_j$ to the prediction f(x) of a ML model is computed using Shapley values by assuming $\{x_1, ..., x_n\}$ as players in a coalition game (27, 28) expressed as (v, N=$\{x_1, ..., x_n\}$). The payoff function v: $2^N \rightarrow R, v(\emptyset) = 0$ maps subset of features (cooperative players) to the real numbers. For a subset of features S, v(S) is equal to the expected sum of payoffs obtained via cooperation of features in S. Once the payoff function is defined, the Shapley value of j-th feature ($\phi_j(v)$) can be computed as the average marginal contribution of the j-th feature to the payout:

$$\phi_j(v) = \sum_{S \subseteq N \setminus \{j\}} w_{S,N}(v(S \cup \{j\}) - v(S))$$

where the summation is computed over all possible coalitions S such that j-th player is excluded. Moreover, $w_{S,N}$ is the weight factor computed as:

$$w_{S,N} = \frac{|S|! \, (|N| - |S| - 1)!}{|N|!} = \binom{|N|}{1, |S|, |N| - |S| - 1}$$

where |S| is the cardinality of subset S and $w_{S,N}$ is equal to the inverse of multinomial coefficient representing the number of different ways of forming coalition using subset S of N excluding j-th feature (i.e. S $\subseteq$ N\{j}). Using Shapley values $\{\phi_j(v), j = 1, ..., n\}$, the SHAP explanation can be computed as:

$$g(z') = \phi_0 + \sum_{j=1}^{n} \phi_j z'_j,$$

where $z' = [z'_1, ..., z'_n] \in \{0,1\}^n$ is a vectors of zeroes and ones. For $z'_j = 0$, j-th feature is not part of the coalition whereas $z'_j = 1$ means j-th feature is present in the coalition. Moreover, $\phi_0 = E[f(x)]$ is the average predicted value computed over all features.

It is not easy to optimize the SHAP method's implementation for each model type, even if it may be used for several models. The high-level ontology of explainable methods to artificial intelligence is given in Figure 3. CNNs often produce class activation maps (CAMs) (29). Per-class weighted linear summation of distinct spatial patterns occurring in an image is indicated by CAMs. Before the output layer, the final convolutional feature map is sent to global average pooling. The input features to a fully connected layer and the output features of a loss function are then produced from this pooled feature map data. By re-projecting the output weights back to the previous convolutional layer, a heatmap depiction of the input image highlights the regions that have a stronger effect on the CNNs' choice. For fully convolutional neural networks, CAMs cannot be used with trained models or those that do not follow the defined architectural guidelines.

CAM and Grad-CAM (30) are based on the assumption that for each specific class c, the final score $f^c$ of the network can be expressed as (31):

$$f^c = \sum_k w_k^c \sum_i \sum_j A_{ij}^k, \tag{1}$$

where $w_k^c$ is the weight corresponding to $A^k$ which is the k-th feature map of the last convolutional layer and $A_{ij}^k$ is the value at row i and column j of $A^k$. Given equation 1, for class c, the saliency map value at each location (i,j) is computed as (31):

$$L_{ij}^c = \sum_k w_k^c A_{ij}^k,$$

where $L_{ij}^c$ reflects the importance of location (i,j) for class c. Therefore, $L^c$ acts as a visual explanation corresponding to class c that has been predicted by the network. In CAM method, weight values $w_k^c$ are estimated by training multiple linear classifiers (one per each class). Moreover, CAM assumes that the penultimate layer of the CNN in question is global average pooling which is not always the case. Grad-CAM was proposed to address the shortcomings of CAM. It has been shown that for each feature $A^k$, weight values $w_k^c$ can be computed as (31):

$$w_k^c = \frac{1}{Z}\sum_i \sum_j \frac{\partial f^c}{\partial A_{ij}^k} \qquad (2)$$

where Z is number of pixels in $A^k$. Despite addressing issues of CAM method, Grad-CAM yet suffers from drawbacks such as failing to properly localize objects in case input image contains multiple instances with identical class labels. Moreover, the averaging in equation 2 is unweighted which leads to localizing only parts of objects instead of their entirety. To overcome these issues, Grad-CAM++ (31) was proposed in which the global averaging in equation 2 is reformulated as (31):

$$w_k^c = \sum_i \sum_j \alpha_{ij}^{kc} . relu\left(\frac{\partial f^c}{\partial A_{ij}^k}\right), \qquad (3)$$

where $\alpha_{ij}^{kc}$ is the weight corresponding to pixel (i,j) of k-th feature map $A^k$. Using equation 3, $w_k^c$ specifically captures the importance of feature map $A^k$ by using weighted average of partial derivatives. This is in contrast to equation 2 in which unweighted averaging is performed. Plugging equation 3 into equation 1, taking derivatives twice with respect to $A_{ij}^k$ on both sides and performing some manipulation yields (31):

$$\alpha_{ij}^{kc} = \frac{\frac{\partial^2 f^c}{(\partial A_{ij}^k)^2}}{2\frac{\partial^2 f^c}{(\partial A_{ij}^k)^2} + \sum_a \sum_b A_{ab}^k \{\frac{\partial^3 f^c}{(\partial A_{ij}^k)^3}\}}$$

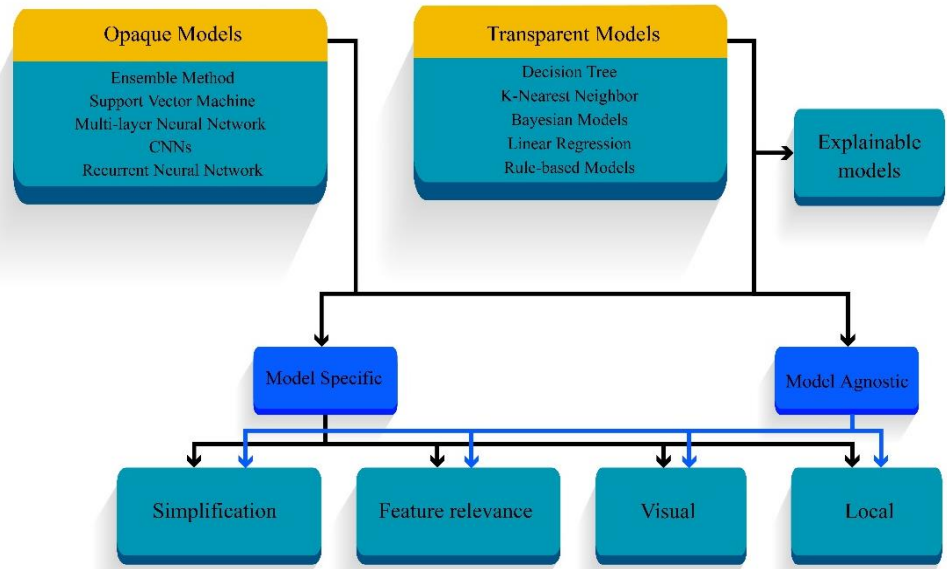

Figure 3. High-level ontology of XAI approaches

Figure 3 demonstrates that while feature-oriented approaches can identify the location of a decision in the input, they do not provide a human-level explanation of the model's reasoning process. To elaborate, while feature-oriented approaches are useful in identifying the specific input features that lead to a decision, they do not provide a complete understanding of how the model arrived at its final judgment. This lack of transparency can be a significant obstacle in certain fields, such as healthcare or finance, where it is crucial to understand model's reasoning procedure. Without a clear and understandable explanation, it would be challenging to trust and effectively utilize the model's output. Therefore, more sophisticated explanation methods are needed to gain users' trust.

## 2.2 Global methods

GAMs (global attribution mappings) may help with explaining neural network predictions across different populations. The benefits of this approach include the ability to record distinct subpopulations at various granularity levels. GAMs discover a pair-wise rank distance matrix between features and groups similar local features using K-medoids clustering. Each cluster's medoid then builds a global attribution based on the patterns found in each cluster. As a result, this approach may feature characteristics across various sample groups. For the majority predicted class, a normalized heatmap depicts the absolute value of the gradient (relative to the input attributes) developed through a gradient-based saliency map. The pixels that have been highlighted show a high level of activation, indicating that they have the most impact (high saliency). Explanation of the method relies on the user's ability to determine what features of an image are being utilized to reach a classification result. However, when propagating nonlinear layers, neurons with negative input gradients are inhibited because of the absolute value, which is the converse of the relative value. Gradient-based saliency maps, like feature-oriented techniques, cannot articulate judgments beyond model diagnosis (32).
.
Deep attribute maps are explored in (33) to increase the explainability of gradient-based algorithms. The model prediction is displayed as a heatmap utilizing the output gradient's important feature, multiplied by relevant input data to compare alternative saliency-based explanation models. The colors red and blue reflect the sound and bad impacts on the final result that each of these components had. The input could create noise and variance in the explanations. Deep attribute maps alone cannot explain why two models yield similar or different outcomes.

## 2.3 Concept models

As pointed out by Kim et al. (34), the complex feature spaces of deep neural networks are not necessarily an obstacle; quite the opposite, deep features can be used to our advantage for model interpretability. Kim et al. have proposed Concept Activation Vectors (CAVs) to represent neural network internal state in human-understandable format. To this end, directional derivatives are used to measure the sensitivity level of neural network output to user-defined concepts. As an example, suppose a neural network is trained to distinguish pictures of horses from zebras. Using CAT, it is possible to measure the contribution level of the animal having body stripes to being classified as zebra.

In a nutshell, a concept which is of interest to the user is represented in terms of multiple examples. The objective is to find a vector in neural network l-th layer activation space that best represents the human-defined concept. To find this vector, in addition to set of concept examples, a set of random examples is prepared. The hyperplane separating activation values corresponding to concept and random examples is determined. The normal vector of the aforementioned hyperplane is considered as the CAV.

## 2.4 Surrogate models

Local interpretable model-agnostic explanation (LIME) constructs locally optimal explanations of ML models using an interpretable surrogate model. While explaining the working mechanism of complex black box models is challenging, it is possible to explain their behaviour for a specific input sample. LIME method starts by modifying parts of the given input sample to generate a dataset of perturbed instances similar to the original input but not exactly the same. The perturbation depends on the nature of the input sample. For example for input of type image, some parts of it can be replaced with gray color to obtain its perturbed counterparts. Figure 4 depicted this process in which the boundaries between image parts (called super-pixels) are shown with yellow color. The black box model is used to predict the probability that each perturbed instance contains a car image. The perturbed instances are weighted according to their predicted probabilities and a linear model is learned using locally weighted regression. The fitted linear model is used to choose the super-pixels with the highest probability of containing a car. The chosen super-pixels are presented as the explanation for the black box model.

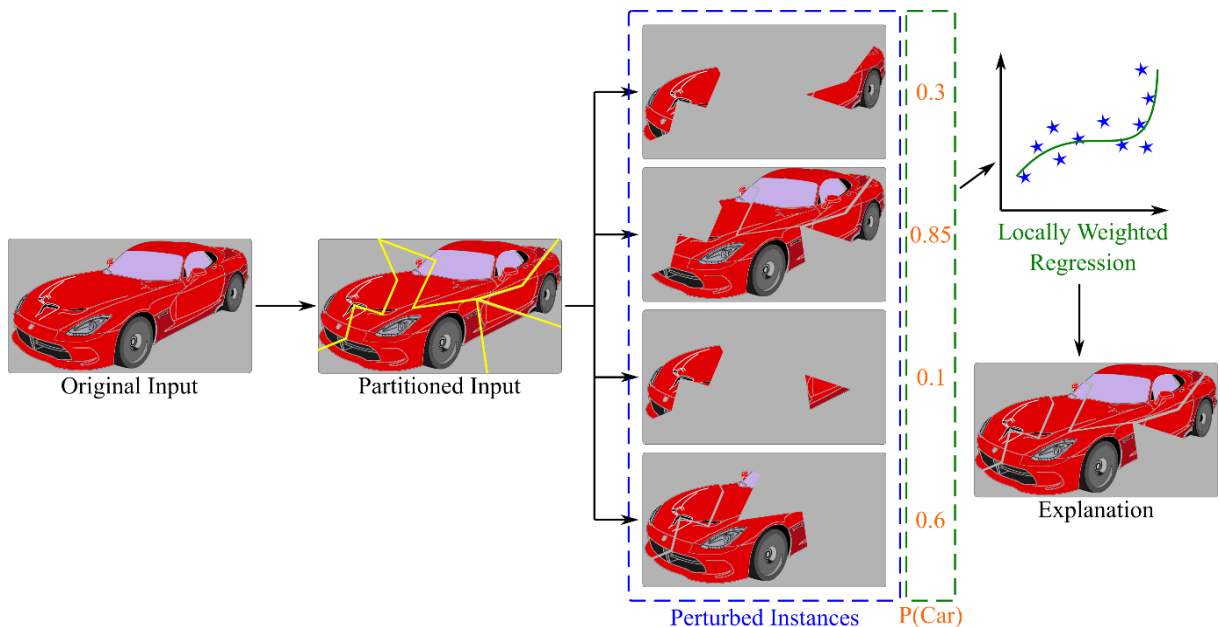
Figure 4. High level steps of LIME method

Figure 4 shows the five high-level steps of the LIME method for generating explanations for individual predictions made by a black-box classifier. As can be seen, the LIME method allows users to generate explanations for individual predictions made by the black-box classifier by creating an interpretable model that approximates the black-box classifier in the vicinity of the chosen data sample.

**2.5 Local, pixel-based methods**

Layer-wise relevance propagation (LRP) employs specified propagation rules to explain a multi-layered neural network's output related to the input. The approach produces a heatmap, offering insight into which pixels contributed to the model's prediction and to what extent. Consequently, LRP emphasizes those variables that positively affect a network's choice. LRP may be applied on an already-trained network to simplify the features' decision-making process if the network employs backpropagation rather than forwards propagation. DeconvNet (35) employs a semantic segmentation approach that constructs a deconvolution network and contributes pixels during the classification process.

**2.6 Human-centric methods**

The previous techniques have failed to offer humans logical explanations despite their benefits. They try to "damage limit" the "black box" by "only touching the surface" by using post-hoc indications about features (attribute allocation) or places within an image. This is in contrast to how people think, develop connections, assess similarities, and draw a comparison. Model structure and parameters, which are critical to the problem's nature, are utterly ignored by the methods described above so is the logic.

Recently, a fundamentally new approach to explainability was proposed (36) which is based on human-centric (anthropomorphic) understanding rather than pure statistics. Humans assess commodities (e.g., photographs, music, and movies) in their totality, not by feature or pixel. People utilize similarity to relate new data to previously learned and aggregated prototypes (32), whereas statistics employ averages as their foundation.

III. **Explainable Artificial Intelligence tools**

To facilitate the behavior analysis of AI models, various tools have been developed. Some of these tools are briefly introduced below:

- ELI5 library (37) developed by MIT is a python library for visualizing and debugging ML models. ELI5 supports various ML frameworks such as Scikit-learn, Keras, LightGBM, etc. The prepared explanations are offered in different data formats such as text, HTML, web dashboard, or JSON.

- AI Fairness 360 from IBM (38) is an open-source library developed to ease the process of detection and alleviation of bias in ML models as well as datasets. This library is available in python and R.
- Interpretability packages from Microsoft (39) offer several model-agnostic/specific explainers (e.g. SHAP tree explainer and SHAP deep explainer) for ML models based on SHAP framework. These packages are available as part of Responsible AI dashboard of Microsoft Azure Machine Learning.
- "What If Tool" (WIT) offered by Google (40) provides visual probing for trained ML models behaviours while reducing required coding as much as possible. WIT can be integrated with several platforms such as Jupyter/Colaboratory/Could AI notebooks, TensorBoard, etc. Moreover, analysis tools for various ML problem categories such as binary/multi-classification and regression are available for wide range of data types such as tabular, image, and text.
- H2O platform offered by H2O.ai (41) aims to accelerate development and deployment of AI models in various business problems. H2O hides the technical details of AI models from users to enable development of AI applications without writing any codes.
- Distill (42) was a research conducted to combine different interpretability techniques to enhance analysis of neural networks decision making process. To this end, Distill exploits the complementary roles of interpretability methods such as feature visualization, attribution, and dimensionality reduction by treating them as building blocks of a unified interface.
- Skater developed by Oracle (43) is a python library for analyzing behavior of trained models both globally and locally. In global scenario, the analysis is done based on inference over a complete dataset. In local scenario, the analysis is performed based on prediction for a single sample.

IV. **Explainable AI for decision makers**

XAI examines if ML models can be more understandable to humans, and aims to enhance their effectiveness as well as making them feasible for non-experts to apply in diverse settings. Despite several attempts made by researchers, yet the exact definitions of concepts such as interpretability and explainability are not readily available. Transparency and XAI popularity has increasing trend in response to resolving the AI "black box" dilemma. A ML model's output may be better understood using XAI methods. Additionally, explainability is related to the explanation as a means of communication between humans and decision-makers in a reliable and comprehensible manner. Development of AI algorithms has an increasing trend for medical applications. These models can aid medical practitioners in offering improved treatment regimens, assisting ailment detection, and monitoring disease progression (44). Sophisticated AI methods may also free up scarce resources by automating tasks that were traditionally handled by specialists like radiologists, whose mobility and adaptability are limited. Moreover, the increasing demand for high-quality medical services can be addressed by AI-based healthcare systems. Yet, the majority of these AI-based models remain prototypes and never make it to market (45). The models were unsuccessful to live up to performance expectations, crucial risky judgments, and depended on differentiating traits to anticipate outcomes. Numerous programs such as Google Health have all failed in some way when placed into test or production (46). This has led to outpouring criticism about AI-based outcomes due to the potentially dire ramifications for humans making high-stakes medical choices. High complexity of the underlying models, large quantity of the datasets, and the huge processing power required to boost the performance of ML models are typical features behind AI models (47). However, as these models get more complex, it becomes more challenging to grasp their operation, data processing, and decision making which is why they are referred to as opaque or black-box models (48). In black-box models, measures such as accuracy, precision, and forecast speed take primacy, impairing the users' capacity to interpret critical outcomes, and relegating them to the abilities of the illiterate. Transparent AI is in great demand because of its utility in high-stakes decision-making such as medical applications. Recent studies suggest that model transparency, interoperability of outcomes, and a clarity of the clinical workflow are all required to ensure that these powerful models are utilized and adapted correctly in healthcare practices.

There has been an attempt to investigate various explainability methods (49). First, approaches were grouped into four categories with the following objectives in mind: 1. describing black-box models, 2. assessing black box models, 3. explaining their outcomes, and 4. building transparent black-box models. Moreover, a taxonomy was proposed to express the underlying explanator, input data type, the issue discovered by the approach, and the "opened" black box model. It has been shown that most explanation methods are unable to decipher models (50). This involves making judgments based on unknown or latent traits. Finally, an explanation is provided for the lack of interpretability techniques in the suggested systems. In addition, an approach was proposed for learning models directly from explanations after recognizing a lack of formality and performance evaluation of interpretability methods. Interpretability framework has also been proposed for developing predictive accuracy, descriptive accuracy, and relevancy (51).

Misztal-Radecka and Indurkhya (52) proposed a taxonomy to classify explainability of DL which includes explaining deep network processing, explaining deep network representation, and explaining how systems are built. Last but not least, establishing transparent or explicable models, fostering cross-disciplinary cooperation between healthcare practitioners is crucial for information sharing (53).

Many scientists believe that XAI make AI deployment easier in the medical industry since XAI helps with creating trust and understanding among stakeholders about AI system (54, 55). According to research on the information required when a difficult model is put into a decision-making environment, the information given by XAI frameworks is of primary importance. For instance, it has been observed that physicians are interested in the local, case-specific logic behind a model choice and the model's global properties (56). XAI frameworks such as LIME, SHAP, or PDP may deliver local and global model information.

Lack of understanding of the model functionality makes it harder to trust the model output (55). Knowing more about how the algorithm was developed would reinforce trust in the AI application's outcomes. However, very little research has been conducted to discover the specific information required when employing AI in algorithm-assisted decision-making in the medical field. These discoveries spurred our research on XAI techniques and their promise to simplify understanding of complex AI models.

Gritzalis et al. (56) showed that explainability is all about offering a relevant explanation for models whereas interpretability refers to the ability to attribute subjective meaning to an object. Interpretation is the process of using one's mental faculties to make sense of numerical data. XAI is where "transparency" first appeared, at a far more technical level. However, there is a key tradeoff between complexity and interpretability when describing XAI products in terms of their usefulness to humans or models. For those who are not acquainted with AI, the difficulty level of understanding the explanation is a major factor. The explanation may be incomprehensible to stakeholders who have no experience with AI, even if it is technically possible to explain the model output or how it generated the result using XAI frameworks.

Došilović et al. (57) proposed a taxonomy for neural network interpretability methods with three separate groups. However, the primary focus was on DL. The first category covers methods that mimic data processing to give insights into the links between the model's inputs and outputs. Methods that explain how data is represented within a network go into the second category, whereas transparent networks that explain themselves fall into the third category. The authors acknowledged that XAI methods are beneficial but also pointed out the lack of combinatorial explanation approaches for integration of diverse explanation methods.

Using a variety of specialized scientific journals, Alonso-Fernandez et al. (58) conducted a comprehensive review of the literature. They underlined the necessity to integrate more formalism into the field of XAI and the interaction between humans and robots. After recognizing the community's predisposition toward examining explainability exclusively through the lens of modelling, they proposed adding explainability into other elements of ML. Finally, they provided a viable research direction which is synthesizing existing explainability methods.

Alonso and Catala (59) have begun to address the call for additional knowledge about the structure of explanations through conceptual papers that have discussed the origins of explanations, how we are biased in our interpretations of explanations, and how explanations are phenomena that occur in the context of human interaction.

The growing research on medical imaging aims to follow the same route as the wider literature on XAI in healthcare. XAI is considered as a sophisticated method that yields universal explanations without consideration of the individual needs of distinct models. As a result of these considerations and the socially constructed notion of explainability, we conducted a case study to determine where the demand for XAI comes from and the specifications of alternative modelling approaches. As such, this research seeks to clarify how XAI influences the development, adoption, and use of AI-based technologies in healthcare.

## V. Applications of Explainable AI in Healthcare

Nowadays, Artificial Intelligence (AI) plays an essential role in pursuing critical systems such as education (60), healthcare (61-70), renewable energy (71), transportation, and traffic (72) that directly impact our daily lives. In the healthcare domain, the applications of AI techniques are in constant progress. However, AI applications and models in healthcare practice require transparency and explainability since inaccurate predictions may have severe consequences (73). Clinicians demand to understand AI systems reasoning as a prerequisite for building trust in the predictions and adoption of AI applications (74). Reliability, accuracy, and transparency (75) are critical requirements, especially for healthcare decision-makers (76). Therefore, AI researchers and practitioners have focused on explaining the decisions made by AI applications such as ML or Deep Learning (DL).

AI algorithms should provide clinicians with understandable explanations about their outputs (77). For example in disease diagnosis, XAI can reveal the features that contribute to AI model output on patient's condition.

To understand the relationship between microbial communities and phenotypes, SHapley Additive explanation (SHAP) algorithm was used (78). The motivation is that SHAP technique can explain the prediction of specific prototype values depending on the outputs of each impactful parameter. The features positively impact predicting the target value if the SHAP values are positive.

Dopaminergic imagery techniques like SPECT DaTscan were analyzed for early diagnosis of Parkinson's Disease (79). The LIME technique was used to accurately classify the Parkinson's disease from the given DaTscan with the appropriate reasoning of the same.

Acute critical illness detection is another important use case of XAI approach in medical research. For example, an early warning score system has been proposed (80). The system was able to explain its prediction with the Electronic Health Record data information using SHAP technique.

XAI for the diagnosis of Glioblastoma based on topological and textual features has been investigated as well (81). The AI model on the fluid-attenuation inversion recovery for the Glioblastoma multiform classification was validated. The local feature relevance to the sample in the test set was computed using LIME method.

A computer-aided system with the capability of explainable sentences has been proposed for lung cancer diagnosis (82). The LIME method was used to develop the local post hoc model, which transforms the critical, relevant feature into natural language. An ensemble clustering-based XAI model was proposed for diagnostic analysis of traumatic brain injury (83). In this model, the expert medical knowledge was combined with the automated data analysis to develop the explainable framework. A model for the COVID-19 detection using the chest X-ray images named COVID-NET has been proposed (84). This model showed 93.3% accuracy and a 91.1% sensitivity on the Covid dataset. The XAI technique GSInquire was used to examine the COVID-NET model's output. To predict the post-stroke hospital discharge disposition, an interpretable ML method has been proposed (85). Linear regression was selected as the baseline model and was compared with the black-box model. The LIME method was used to identify the essential features of the model. For selecting the laser surgery option at an expert level, a multicast XGBoost model has been proposed (86). The model was validated on the subject who has undergone refractive surgery. An accuracy of 78.9% was achieved on the external validation dataset. The SHAP method was used to provide a clinical understanding of the ML method.

Ye et al. proposed (87) classification models for COVID-19 equipped with XAI strategies to deliver reliable classification associated with credible explanations for COVID-19. To this end, 380 positive (having COVID-19) and 424 negative CT volumes were obtained. Their model can assist radiologists in determining exact location of lesions in CT scans by providing more diagnostic information. They compared their proposed XAI modules with other models like CAM, SHAP, etc.

Another work (88) has presented a B5G architecture for detecting COVID-19 utilizing CT scan images. The inspection system was developed based on different functionalities of 5G networks. XAI model was utilized to monitor mask-wearing, social distancing, and body temperature. The suggested healthcare framework utilizes three layers which are edge layer, stakeholder layer, and cloud layer. In the middle edge layer, the Local Interpretable Model-Agnostic (LIMA) model is used by the XAI module Knowledge mapping. Images from X-rays, CT scans, and ultrasounds, along with learnable parameters from different layers of the DL model form the input. Their methodology aids with the reduction of hospital overcrowding, the verification of non-COVID-19 patients, and the processing of sensitive personal data at the edge to protect anonymity.

El-Sappagh et al. (89) proposed a model for identification and progress prediction of Alzheimer's disease (AD). The approach consists of two layers. The first layer utilizes random forest (RF) to classify patients who have AD in its early stage. The second layer of binary classification is carried out to predict likelihood of mild cognitive impairment (MCI) progression toward AD within three years from the initial diagnosis. In each layer, SHAP is used for providing explanation on the model reasoning. The evaluation of the proposed approach was done on ADNI dataset achieving high reliability of 93.95% and 87.08% per layer. Sepsis must be diagnosed as soon as possible since delayed treatment leads to patient's irreversible organ damage increasing the mortality rate. Yang et al. (90) tackled early diagnosis of Sepsis based on health records obtained from Cardiology Challenge 2019. Using 168 features collected on hourly basis, an explainable AI model was developed for sepsis diagnosis. Gradient-boosting-trees model called XGBoots was used in K-fold cross-validation setup to forecast sepsis and provide interpretable sepsis risk in the ICU.

Chakraborty et al. (91) have used XAI to investigate the relationship between tumor immune cell composition and breast cancer survival rates. Using EPIC, TIMER, CIBERSORT and xCell computational approaches, from TIMER2.0 and TCGA breast invasive cancer data, first they extracted immune cell from a RNA bulk sequencing data. According to the proposed XAI techniques the most significant cells responsible for breast cancer are the M0 macrophages, B cells, CD8+, and NK T cells. Their model demonstrated that increasing the fraction of B cell with CD8+ T and NK T cell, their points of inflection might increase survival rate of breast cancer patients up to 18%.

Dave et al. (92) have studied and exploited different XAI methods in the healthcare sector. LIME and SHAP feature-based techniques have been applied to a heart disease dataset with 70+ features taken from UCI ML Repository. Additionally, they have compared and discussed other popular techniques like Anchors, Contrastive Explanation Methods, Counterfactuals, Integrated gradients Kernel Shapley, etc. The primary goal was to investigate how these strategies are beneficial and how certain aspects are accountable for the model's outcomes.

Congenital heart disease (CHD) diagnosis using fetal ultrasound screening faces multiple challenges such as technical differences between examiners and manual operation. To address these challenges, CNN was utilized for detection of cardiac substructures and structural abnormalities in fetal ultrasound videos leading to a new method called SONO (93). The detection probability was expressed using a barcode-like timeline used as one of the features of XAI. The proposed method was evaluated on 104 sets of 20 sequential video frames. The sequences consist of cross-sections around three-vessel trachea view (3VTV) (Vessels) and around 4CV (Heart). The sequences were obtained from 40 normal and 14 CHD cases. It has been claimed that SONO performed better than other anomaly detection algorithms and achieved a high detection rate in significant substructures, whereas relatively small substructures like pulmonary vein, tricuspid valve, etc. were untraceable.

The work done by (94) attempted to assess the efficacy of several deep-learning algorithms in locating tumour tissues and separating them from healthy areas in the brain. TCGA dataset was used for conducting experiment, consisting of 34,800 slices of brain images. Three deep networks namely DenseNet-121, GoogLeNet, and MobileNet were evaluated in the experiments. Reported results showed that DenseNet-121 achieves better tumor localization with ~80% hits.

Varzandian et al. (95) have adopted Magnetic Resonance Imaging (MRI) scan of brain with 1901 different subjects obtained from IXI, ADNI and AIBL repositories. To categorize patients suffering from AD, they trained an analytical model constructed on chronological and brain age data. They have argued that this model offers superior performance compared to other ML methods for females and males with 88% and 92% accuracy. The authors developed a methodology for performing regression and classification tasks while retaining the input space's of morphological semantics and giving a feature score to quantify each morphological region's detailed contribution to the final outcome.

Apart from importance of interpretability, Tanzeela et al. (96) have recognized the necessity of ethics of AI by presenting a survey on ethical solutions for deployment of AI in different application domains. The survey points out some concerns regarding AI taking over our lives. For example, while automation using AI is beneficial in terms of lowering the production cost, it may lead to unemployment in human workforce. Moreover, companies that utilize AI will receive much higher profits faster compared to companies running on human workforce. This is unfair for companies that cannot afford AI automation. The authors also discuss the details of preparing high quality data which is necessary for training robust and reliable AI models.

The authors present a system in (97) for content-based image retrieval (CBIR) of Video frames related to minimally invasive surgery (MIS) videos. In the proposed method, descriptors were extracted that were semantic in nature from mentioned video frames.

In Case-Based Reasoning (CBR), a pool of labelled samples is available and labelling a new query sample is done based on the labels of its similar counterparts fetched from the pool. Contrary to DL, CBR reasoning is clear since it is primarily based on some type of similarity measure between labelled samples and the query sample. Therefore, Lamy et al. (98) proposed a CBR method for breast cancer diagnosis equipped with a user interface for providing visual explanations.

Prostate cancer is very common among men worldwide and its early diagnosis is vital to patient's survival chance. Hassan et al. (99) took some sort of hybrid approach to boost performance of prostate cancer diagnosis based on ultrasound and MRI data. In this approach, multiple pre-trained DL models are fused with classic methods such as RF, SVM, etc. To make the results interpretable, LIME approach was utilized.

It is beneficial to close this section by summarizing XAI research papers related to medicine and healthcare in Table 2. The applied ML and XAI approaches as well as studied diseases are listed for each paper.

Table 2. Summary of various XAI methods in digital healthcare and medicine, including their ML and XAI Methods

| Disease/ Type of Images or documents | ML Methods | XAI Methods | Refs | Year |
|---|---|---|---|---|
| Spine | One-class SVM, binary RF | LIME | (100) | 2021 |
| Glaucoma | EAMNet based on CNN | Visual, Post-hoc | (101) | 2019 |
| Skin Cancer | CNN, Random Forest, KNN | A visual, naturally interpretable model | (102) | 2021 |
| Histology and radiology images | StyleGAN | Visual, Local, heatmap-based interpretability | (103) | 2021 |
| EHR | RNN | Post-hoc, local, decision tree | (104) | 2020 |
| Retinal fundus images | ResNet50 | SIDU, GRAD-CAM | (105) | 2021 |
| Autism Spectrum Disorder (ASD) | SVM | feature importance score, Visual | (106) | 2021 |
| Hepatitis | LR, DT, kNN, SVM, RF | SHAP, LIME, partial dependence plots (PDP) | (107) | 2021 |
| Traumatic brain injury (TBI) identification | k-means, spectral clustering, Gaussian mixture | Quality assessment of The clustering features. | (108) | 2020 |
| Colorectal cancer diagnosis | CNN | Visual explanation | (109) | 2020 |
| Automatic recognition of instruments in laparoscopy videos | CNN | Activation Maps | (110) | 2019 |
| Decision Support System for Prostate Cancer | Gradient-boosting algorithm | SHAP | (111) | 2020 |
| ECG Based Hypoglycaemia | PCA | Grad-Cam | (112) | 2020 |
| Inflammatory bowel disease diagnostic | Linear Support Vector Machine (SVM), Random Forest (RF), Nearest Shrunken Centroids (NSC), and Logistic Regression with L2 regularization (LR). | Feature Marginalization | (113) | 2017 |
| Macromolecular Complexes | CNN | ML-CAM | (114) | 2018 |
| Allergy diagnosis | Decision Tree, Support Vector Machine and Random Forest | Post-hoc XAI and CDSS | (115) | 2021 |
| Glaucoma Diagnosis | CNN | ML-CAM | (116) | 2022 |
| Pneumonia identification | VGG16 | Grad-CAM | (117) | 2022 |
| Alzheimer's disease | VGG-16 and CNN | Feature importance score, Visual | (118) | 2022 |

## VI. The Challenge of Interpretability in Healthcare

In this section, we look into the challenge of XAI in healthcare. We discuss the barriers that prevents the widespread application of XAI in healthcare and medicine.

### 6.1 User-Centric Explanations

Understanding internal process of ML methods and outcomes in healthcare systems is essential to make such critical systems more trustworthy for end-users (patients and clinicians) (119-122). The challenges that emerge in making ML models explainable are listed below (123, 124) :
1. Analysis of complex ML models requires background in advanced mathematics and statistics.
2. So far, healthcare systems have failed to fulfil the design and functional requirements for successful deployment of ML models in medical domain (125, 126).
3. The end-users' desire towards interpretability widens the gap between development of complex black box models and human-readable explanations (125).
4. Making ML models more transparent is likely to make them less efficient in achieving their objectives. This is due to the fact that high performance models consists of many layers with complex interconnections. Therefore, tracing the training process on millions of samples is almost impossible (125).
5. XAI methods only highlight the regions relevant to the ML model outputs without determining the features that have caused the relevancy of those regions (127).
6. While considerable effort has been put into making models transparent, the appropriate evaluation of provided explanations is still an open issue (128). Moreover, some researchers have doubts about reliability of XAI methods considering them to be misleading (129).

Several studies (126, 130-133) have shown that for delivering accurate and trustworthy explanations and predictions, it is critical to focus on end-users. As a result, end-users must be involved in developing ML models to bridge the gap between user needs and expectations and design support of the products. In applications such as detecting objects (e.g. vehicles, animals), recognizing actions, controlling robots, etc. lack of user participation may not be much of an issue. This is because ML experts can analyse XAI methods outputs to debug models and determine training data gaps on their own. However, in medical domain, the situation is different. Even if XAI methods provide plausible explanations, only clinicians can analyse XAI outputs and understand the cause of fail cases for ML models. Therefore, ML experts always have to rely on clinicians for debugging and improving their models. Considering that clinicians are usually busy with their own tasks, collaborating with them would be challenging. Similar concern has been recognized by Brujin et al. (134).

Another challenge encountered in applying XAI in medical domain is the fact that ML experts are usually comfortable with mathematical explanation outputs. On the contrary, clinicians prefer to receive explanations in visual form (127). Such requirement puts limitations on the output format of XAI methods.

### 6.2 Performance vs. Transparency Tradeoff

XAI in healthcare systems impacts how end-users comprehend ML models decisions. Therefore, it is essential to balance the trade-off between the model's complexity and accuracy. Explainability is inversely related to performance of AI systems. Increasing the model transparency improves the ability to analyze its decisions (4, 135, 136). Consequently, the XAI models divide into black-box AI, grey box AI, and white box AI. DL and ensembles approaches are included in the black box, statistical models are included in the grey box, and graphical models, linear models, rule-based models, and decision trees are included in the white box (137, 138). The black box of AI in healthcare systems is not transparent, making it difficult to provide acceptable reasoning for fair decisions and end-user trustworthiness (4, 136, 139). The gray-box AI maintains average balance between transparency and explainability, while the white-box AI has high explainability with low-performance models. Ideally, models with high explainability and acceptable performance are desired for healthcare systems. However, there is a trade-off between the ability to discover understandable patterns and flexibility in fitting the data with high accuracy (101). The trade-off must also be discussed with end-users to grasp the clinical and human risks associated with misclassification.

### 6.3 Balancing Requirements of Interpretability

Considering that interpretability is a complicated and nuanced term with no single definition, several requirements for an ideal interpretable ML system should be stated specially for healthcare applications. In general, the ML explanation is related to model's soundness (or optimality) and ability to be comprehended by the user (136, 140). In addition to soundness and comprehension, it is crucial to cover the explanation scope of models from a local (or instance-based) or global level. The global level reduces model performance, while the local level increases the time complexity to provide a comprehensive explanation (124, 141). The challenge is to ensure soundness, comprehension, and scope of the model and satisfy trustworthiness about the black and grey box AI model working mechanisms. The soundness and comprehension requirements are balanced based on the sensitivity of the application domain and the amount of which the end-user is anticipated to identify the ML model's interpretability.

### 6.4 Assistive Intelligence

The ultimate objective of ML algorithms is removal of humans from decision making in various application domains (142). However, in safety critical applications such as healthcare, the decision making cannot be left to ML systems entirely. Supervision of human experts is necessary to avoid catastrophe in case wrong decisions are made by the ML system. While ML methods cannot be fully trusted with patients' lives, they can act as medical assistants for human experts accelerating medical data analysis and useful knowledge extraction. Healthcare systems require precise data to make robust decisions so human-in-the-loop framework as well as XAI mechanisms are needed.

## VII Conclusion

AI has induced a significant paradigm shift in the healthcare sector and has revolutionized data access and analytical methods. The introduction of XAI has further accelerated the development of different AI techniques. In this paper, we reviewed the literature of XAI with special focus on healthcare applications. We categorized XAI methods based on the functionality and algorithmic procedure and provided a survey including 26 medical diagnosis and surgery papers using explainable artificial intelligence. The challenges of applying XAI healthcare domain were presented. Moreover, we alluded to the crucial role of model explainability for the domain of healthcare and the importance of human-in-the-loop in design and development of XAI methods in order to interpret AI models for patients and medical experts. In recent years, XAI has played a profound and noteworthy part in healthcare which includes both diagnosis and surgery applications. Yet, there is a lack of integrated explainability tools to incorporate XAI with ML and DL methods to perform diagnosis and treatment suggestions, especially in surgery and other critical medical procedures. XAI has a high impact on healthcare because of the increased demand for trustworthy and transparent AI models from medical professionals. To this end, the collaboration of data scientists and medical experts for design and development of more efficient XAI applications will be certainly required. Such applications can affect the medical procedure and treatment in a significant manner and increase the satisfactory level of patients by understanding the factors that cause diseases and measuring the influence of each types of medications.

Conflict of Interest: There is no conflict of interest.